\documentclass[final,11p,times,twocolumn]{elsarticle}

\usepackage[top=2.5cm,bottom=2.5cm,left=2.0cm,right=2.0cm]{geometry}
\usepackage[labelsep=period]{caption}
\usepackage[figurename=Fig.]{caption}
\usepackage{subcaption}
\usepackage{booktabs} 
\usepackage{comment}
\usepackage{mathtools}
\usepackage{textcomp}
\usepackage{gensymb} 

\usepackage{amssymb}

\usepackage{lineno,hyperref}
\DeclareMathAlphabet{\mathcal}{OMS}{cmsy}{m}{n}
\DeclareMathOperator*{\argmin}{arg\,min}

\journal{ICASP14}

\begin{document}

\begin{frontmatter}



\title{Active learning for structural reliability analysis with multiple limit state functions through variance-enhanced PC-Kriging surrogate models} 


\author[1]{J. Morán A.\corref{cor1}}
\ead{jmoran@uliege.be}
\author[2]{P.G. Morato}
\author[1]{P. Rigo}

\cortext[cor1]{Corresponding author}
\address[1]{ANAST, Department of ArGEnCo, University of Liege, 4000, Liege, Belgium}
\address[2]{Department of Wind and Energy Systems, Technical University of Denmark, 4000 Roskilde, Denmark}





\begin{abstract}
Existing active strategies for training surrogate models yield accurate structural reliability estimates by aiming at design space regions in the vicinity of a specified limit state function. In many practical engineering applications, various damage conditions, e.g. repair, failure, should be probabilistically characterized, thus demanding the estimation of multiple performance functions. In this work, we investigate the capability of active learning approaches for efficiently selecting training samples under a limited computational budget while still preserving the accuracy associated with multiple surrogated limit states. Specifically, PC-Kriging-based surrogate models are actively trained considering a variance correction derived from leave-one-out cross-validation error information, whereas the sequential learning scheme relies on U-function-derived metrics. The proposed active learning approaches are tested in a highly nonlinear structural reliability setting, whereas in a more practical application, failure and repair events are stochastically predicted in the aftermath of a ship collision against an offshore wind substructure. The results show that a balanced computational budget administration can be effectively achieved by successively targeting the specified multiple limit state functions within a unified active learning scheme.
\end{abstract}


\begin{keyword}
Active learning; Gaussian Processes; Structural reliability; Surrogate modeling; Offshore wind turbines; Polynomial chaos expansion.
\end{keyword}

\end{frontmatter}

\section{INTRODUCTION}
\label{sec:intro}
An optimized design and management of engineering systems from a life-cycle perspective aims at jointly minimizing maintenance costs and structural failure risk, quantified via economic and structural reliability metrics \citep{MORATO2022102140}. To probabilistically characterize a failure event and/or a specific damage condition, one or multiple performance functions can be accordingly formulated, thereby accounting for the uncertainty associated with system response predictions and informing asset management decisions. The quantification of uncertainties associated with specifically defined failure and/or damage events normally demands the computation of several high-fidelity engineering simulations, e.g., finite element analysis, computational fluid dynamics. 

Considering that high-fidelity simulations can be time-consuming and computationally expensive, surrogate models offer an attractive solution by providing a light-running approximation of the model response based on a reduced number of data points. In general, surrogate (meta)models are able to efficiently learn the mathematical relationship between uncertain input design variables and a relevant output quantity of interest (QoI). Among various proposed surrogate models, Gaussian Process-based models (GP), e.g., Kriging and PC-Kriging, have demonstrated their effectiveness in supporting a wide range of engineering applications \citep{Schobi_PCKriging}. Leveraging on their probabilistic formulation, GP-based surrogate models yield not only a point estimate of the QoI but also an uncertainty measure associated with the generated prediction. In many applications, surrogate models seek the minimization of the system response generalization error with the least number of high-fidelity model evaluations, often inducing a global exploration of the design space. When dealing with structural reliability applications, the relevant limit state(s) can be directly surrogated, and the experimental design logically focuses on regions near the boundary in order to accurately estimate the probability associated with a specific event, e.g., failure, damage condition. 

With the goal of efficiently improving the accuracy of a surrogated limit state function, active learning is a machine learning technique that sequentially selects the training samples based on a specified learning metric. This approach is particularly useful when the computational cost of retrieving new high-fidelity model evaluations is high and/or in settings under a constrained computational/economic budget. From the original development of the Expected Feasibility Function (EFF) \citep{doi:10.2514/1.34321} and the U-function \citep{ECHARD2011145}, sophisticated active learning schemes have been widely proposed in the literature, in which the training samples are collected based on an exploitation-exploration trade-off, i.e., exploiting observations near the limit state or exploring yet uncertain ones. As suggested by \cite{MOUSTAPHA2022102174}, active learning approaches can be generally categorized according to four distinctive features: (i) surrogate model choice, (ii) failure probability estimation technique, (iii) learning enhancement metric, and (iv) stopping criteria.

Active learning methods have been proven to be effective for a myriad of structural reliability problems \citep{MOUSTAPHA2022102174}. The computational budget, however, is there mainly dedicated to efficiently capturing a limit state and its associated failure event. In many engineering applications (e.g., inspection and maintenance planning), additional events rather than structural failure may be relevant, hence potentially demanding the evaluation of multiple limit states that inform, for instance, operational decisions. In this work, we investigate the capability of active learning approaches to identify training samples when dealing with multiple interrelated limit states under a common computational budget, in which, an observation can be informative for estimating either the event associated with only one or multiple events. Specifically, PC-Kriging-based surrogate models are actively trained while additionally accounting for a variance correction derived from leave-one-out cross validation error information, whereas the learning scheme relies on sequentially Monte Carlo samples evaluated according to a U-function.  

In order to effectively balance the generation of training points around multiple limit states within a common budget, we additionally propose here active learning strategies that sequentially select observations with the objective of jointly improving the accuracy of all treated limit states. Besides the traditional exploitation-exploration trade-off, training samples are sequentially picked by balancing the predicted accuracy among the surrogated limit states. The proposed active training approaches are then tested in a characteristic multi-modal structural reliability setting, examining their accuracy and training efficiency; and in a more practical application, we efficiently surrogate limit states associated with both failure and repair events corresponding to the aftermath of a ship collision against an offshore wind substructure.

\section{LEARNING LIMIT STATE FUNCTIONS THROUGH SURROGATE MODELS}
\subsection{PC-Kriging surrogate models}

Consider the response of a system represented by $y\in \mathbb{R} $, as a one-dimensional output space which is retrieved from the deterministic mapping of the M-dimensional stochastic input parameter space, whose realizations are denoted as $\mathbf{x} = \{x_1, \ldots, x_M \}^T \in \mathcal{D}_x \subset \mathbb{R}^M$. Within this frame of reference, PC-Kriging (PCK) is a non-intrusive surrogate modeling method that combines the probabilistic features of Gaussian processes, also known as Kriging, with a prior trend specifically tailored to the underlying input random vector, $\mathbf{x}$, via Polynomial Chaos Expansion (PCE). While Kriging estimates the analyzed quantity of interest (QoI), $y$, through a stationary spatial random process where the covariance function (kernel) is defined as a function of the relative distance between data points, PCE further incorporates prior knowledge to the Gaussian process mean, as the sum of orthogonal polynomials described by the joint probability density function $f_{\mathbf{X}}$. Formally, a PCK metamodel, $\mathcal{M}^{PCK}$, is defined as: \citep{Schobi_PCKriging}
\begin{equation}
\mathit{y} \approx \mathcal{M}^{PCK} (\mathbf{x}) = \sum_{\alpha \in \mathcal{A} } a_\alpha \psi_\alpha (\mathbf{x}) + 
\sigma^2 Z( \mathbf{x}),
\label{eq:PCK_eq}
\end{equation}
where the left-hand term corresponds to the mean of a Gaussian process, defined as the linear combination of coefficients, $a_\alpha$, and multivariate orthogonal polynomials, $\psi_\alpha (\mathbf{x})$, specified according to the input random vector, $\mathbf{x}$. The terms of the polynomial sum are multi-indexed through $\alpha \in \mathcal{A} \subset \mathbb{N}^M$. The constant variance, $ \sigma^2 $, is formulated on the right-hand of the equation, along with the zero mean, unit variance, stationary Gaussian process, $Z(\mathbf{x})$. The latter is described by an autocorrelation function $ \mathbf{R} (|\mathbf{x} - \mathbf{x'}|; \theta ) $, where the covariance is usually defined based on the relative distance of its inputs $\mathbf{x}, \mathbf{x'} $, and is parameterized by the hyperparameter $\theta\in \mathbb{R^{+}}$, commonly referred as scale parameter or characteristic length. 



To calibrate a PCK model built from a set of polynomials truncated by the size of $\mathcal{A}$, one can rely on non-intrusive methods that are based on discrete system responses, gathered from a sampling plan of $N$ input realizations $\mathcal{X } = \{\mathcal{X}^{(1)}, \ldots, \mathcal{X}^{(N)}\}^T \in \mathcal{D}_x$, also known as \textit{experimental design}. The parameters $\sigma^2$ and $a_\alpha$ can be calculated through maximum-likelihood, i.e., maximizing the likelihood associated with model predictions, $\mathcal{M}^{PCK}(\mathcal{X})$.
Since $a_\alpha$ and $\sigma^2$ are defined as a function of $\theta$, an optimization process for determining $\theta$ can be formulated as:
\begin{equation}
\hat{\theta} =  \argmin_{\theta \in \mathcal{D}_\theta }  \frac{1}{2} \left[ \log \left(\det \mathbf{R} \right) + N \log (2 \pi \sigma^2) + N   \right].    
\label{eq:PCK_theta}
\end{equation}
In an iterative process, various multivariate orthogonal polynomials of increasing truncation order can be trained following the optimization process described before and further classified according to a specified error metric. For example, the optimal PCK model can be defined as the one that yields the minimum \emph{leave-one-out} (LOO) cross-validated error, $\epsilon_{LOO}^{PCK}$, formulated as:
\begin{equation}
\epsilon_{LOO}^{PCK} = \frac{1}{N}  \sum_{s=1}^{N} \left[ y^{(s)} - \mathcal{M}_{(-s)}^{PCK}\left(\mathcal{X}^{(s)} \right) \right]^2.
\label{eq:LOO}
\end{equation}
This error metric is an estimate of the generalization error based on the current experimental design, which is the mean squared error between the responses $y^{(s)}$ and the model predictions at $\mathcal{X}^{(s)}$ of a PCK model, $ \mathcal{M}_{(-s)}^{PCK} $, calibrated with $\mathcal{X}^{(-s)}$ sample points.

\subsection{Active learning for structural reliability analysis}
Compared with conventional design of experiments, e.g., Latin hypercube sampling (LHS), more accurate predictions can be achieved with less high-fidelity model evaluations if an active training approach is followed, in which a surrogate model is sequentially trained based on the available observations up to that point. Each subsequent experimental design point is selected as the one that results in a maximum expected improvement in model accuracy according to a metric computed from a learning scoring function. Very often, this enhancement metric is mainly driven by the surrogate model's built-in probabilistic features, e.g., variance prediction.

A well-known scoring function is the one proposed by \cite{ECHARD2011145}, in which a metric denoted as `U’ is evaluated for a set of randomly generated samples. A tailored and efficient exploration of the design space can be additionally accomplished if the samples are generated via simulation-based methods, e.g., Monte Carlo, where the evaluated points are directly sampled from the underlying input random variables. More specifically, the above-mentioned U-function focuses on the design subspace near the limit state boundary considering an exploitation-exploration trade-off, as mentioned in Section \ref{sec:intro}. 
Mathematically, each subsequent experimental design point, $\mathcal{X}_g$, is selected as:

\begin{equation}
 \mathcal{X}_g = \argmin_{ \mathbf{x} } U (\mathbf{x})  = \argmin_{ \mathbf{x}} \frac{| \widehat{g}(\mathbf{x})|}{\sigma_{\widehat{g}}(\mathbf{x})}  ,
\label{eq:u_func}
\end{equation}
where $\widehat{g}(\mathbf{x})$ and $\sigma_{\widehat{g}}(\mathbf{x})$ represent the mean and standard deviation, respectively, predicted via a surrogate model. The expected improvement metric favors design points close to the limit state boundary and high associated predicted variance. In order to calculate more accurate expected improvement scores, the variance estimate provided by a PCK model for a specific design point can be corrected \citep{XIAO2018404}. 

The selection of subsequent experimental design points can thus be further improved by adequately correcting the generated variance predictions, $ \sigma_{\widehat{g}}(\mathbf{x})$, which might be potentially biased \citep{le2015cokriging}. To do that, a correction factor can be calculated from a leave-one-out (LOO) cross-validated analysis, which is then applied to the variance predictions associated with the experimental design, $\mathcal{X}_{i}$, within the Voronoi domain $\left( V_{i} \right)_{i=1,...,N}$: 
\begin{equation}
\begin{split}
\sigma^2_{\widehat{g}}(x)_{LOO} = \sigma^2_{\widehat{g}}(x) \left( 1 + \sum_{i=1}^{N} \frac{ \left[ \mathbf{e}^2_{LOO} \right]_{i} }{ \left[ \mathbf{s}^2_{LOO} \right]_{i} } \mathbb{I}_{x\in V_{i} } \right), 
\label{eq:Var_LOOCV}
\end{split}
\end{equation}
where $\mathbf{e}^2_{LOO}$ denotes the vector of LOO squared errors,  $\mathbf{s}^2_{LOO}$ the vector of LOO variances and $\mathbb{I}_{x\in V_{i}}$ the domain of the Voronoi cells $\left( V_{i} \right)_{i=1,...,N}$.

\subsection{Active learning strategies for multiple limit state functions settings}
Active learning strategies developed for reliability analysis sequentially select experimental design points with the objective of more accurately representing a specific limit state function. As mentioned in Section \ref{sec:intro}, multiple limit state functions defined within a common input design domain can be of interest in many practical applications, e.g., informing operational decisions or estimating the probability associated with a structural failure event. If the experimental points are specific to a particular limit state function, $g_j$, predictions generated for another limit state function, $\widehat{g}_k$ might not be necessarily accurate.   

To equally balance a certain given computational budget, a logical active training scheme could, for instance, alternate its focus between $m$ considered limit state functions, improving one at each consecutive step. If the experimental design points, $\mathcal{X}$, are scored via the U-function (Eq. \ref{eq:u_func}), the training sequence will subsequently evaluate the U-function defined with respect to an alternate limit state function from all considered ones, i.e., $\mathcal{X}_{g_{j}}$ with $j=1,...,m$.

In practice, some limit state functions are substantially easier to train, hence rendering accurate predictions when trained with only a few observations. By implementing the alternate active learning strategy, $ \mathcal{X}_{g_{j}}$, the computational budget is equally allocated for all limit state functions, even if a certain limit state is accurately predicted by the surrogate model early in the sequential process. Instead, one can detect which surrogated limit state function is still far from reaching convergence and select the subsequent experimental design point by evaluating a learning scoring function with respect to the identified limit state. At each active learning step, $t$, a convergence-related metric $\delta (\widehat{\beta}_{g_j})$ inspired on the reliability index is here proposed for selecting the target limit state function:
\begin{equation}
\delta (\widehat{\beta}_{g_j})  = \left| \frac{ \widehat{\beta}_{t+1,g_j} - \widehat{\beta}_{t,g_j} }{ \widehat{\beta}_{t,g_j}  } \right|.
\label{eq:beta_strat}
\end{equation}
At each training step, the target limit state function, $g^*_j $, becomes the one associated with the maximum $\delta (\widehat{\beta}_{g_j})$, and hence the experimental design point, $\mathcal{X}_{g^*_j}$, is accordingly chosen from a learning metric (e.g., U-function) estimated according to the identified limit state function.
Note that the index $\beta_{g_j}(\mathbf{x})$ follows an inverse relationship with the probability associated with the event of interest, p$_{g_j}$, commonly defined in the standard normal space as $\beta_{g_j}(\mathbf{x}) = - \Phi^{-1} \{ \mathrm{p}_{g_j} (\mathbf{x})\}  $, where $\mathrm{p}_{g_j}$ is computed as the probability of the limit state being negative, i.e., $\mathrm{p}_{g_j} = \mathrm{p} \{ g_j \leq  0  \}$. 

\section{NUMERICAL EXPERIMENTS}
\begin{figure*}[h]
  \graphicspath{ {./figures/} }
  \centering
  \includegraphics{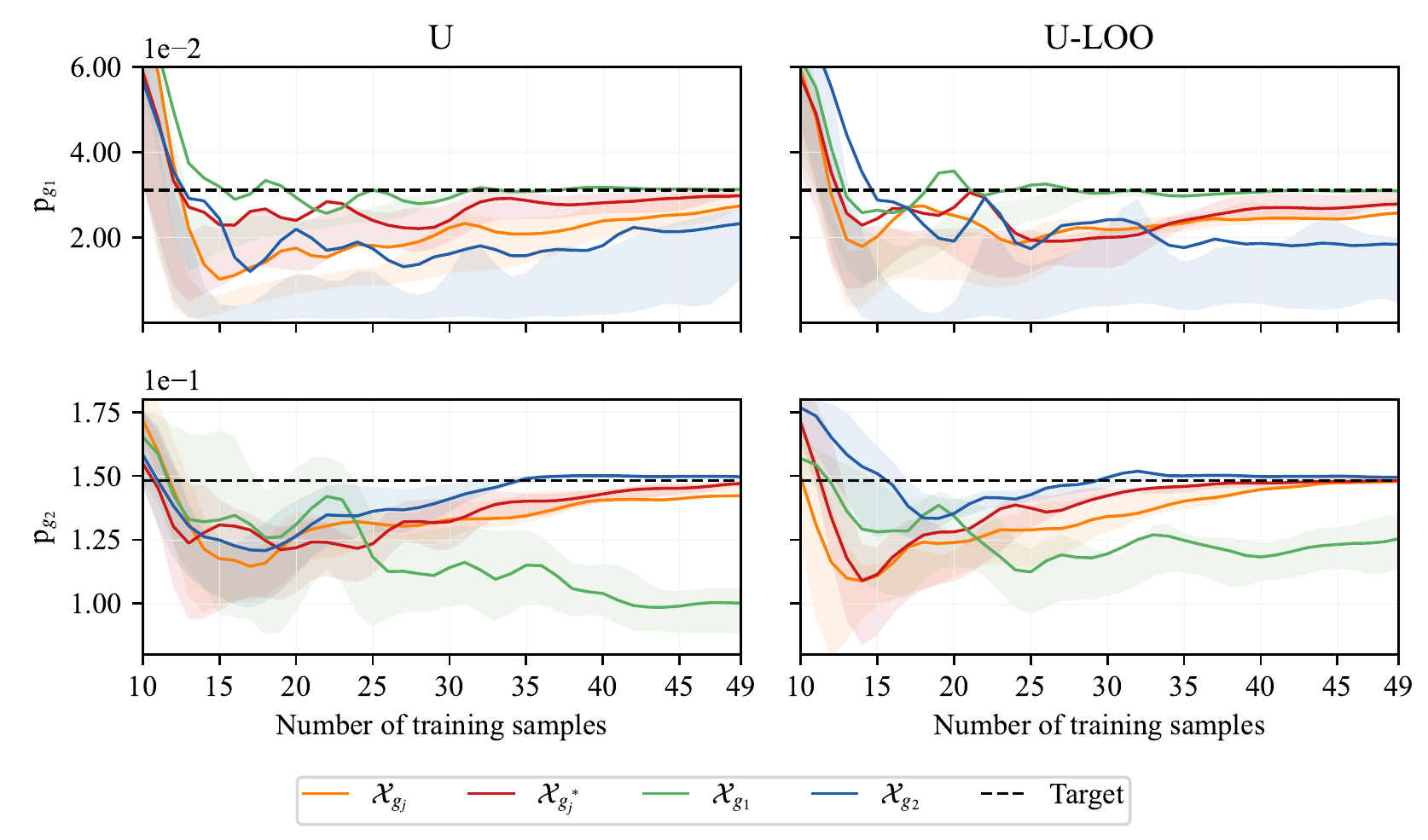}
  \caption{Active learning evolution corresponding to the tested strategies in terms of the predicted probability associated with both considered events $g_1$ and $g_2$. (Left) U-function score metrics evaluated from PC-Kriging variance predictors. (Right) U-function score metrics calculated from corrected variance predictors (U-LOO).}
  \label{Case1: Pf_evolution}
\end{figure*}

\subsection{Analytical reliability problem}
Inspired by \cite{XIAO2018404}, the first limit state, $g_1$, studied here is defined as:
\begin{equation}
\begin{split}
g_1 \left(x_1, x_2\right) = \sin{ \left( \frac{5 x_1}{2 } \right) }  + 2 -  \frac{ \left(x_1^2 + 4\right) \left( x_2 -1\right) }{20}. 
\label{eq:g1_simplified}
\end{split}
\end{equation}
For the sake of capturing the active training performance over a second limit state function, an additional performance function $g_2$ is proposed and formulated as:

\begin{equation}
\begin{split}
g_2 \left(x_1, x_2\right) = \sin{ \left( 2 x_1 \right) }  - \frac{1}{2} +  \frac{ \left(x_1^2 + 4\right) \left( x_2 +1\right) }{20}, 
\label{eq:g2_simplified}
\end{split}
\end{equation}
with random variables, $x_1$ and $x_2$, described as $x_1 \backsim  \mathcal{N} \left(\mu = 1.5, \sigma = 1 \right)$ and $x_2 \backsim  \mathcal{N} \left(\mu = 2.5, \sigma = 1 \right)$, respectively. The tested active learning strategies are restricted to a computational budget of 49 \emph{ground-truth} observations. From the available budget, 10 training points are dedicated to an initial global exploration following a uniform stratification of samples through LHS, committing the remaining resources to the active training of the surrogate model. From the stationary correlation families, the general form of Matérn kernel of degree 5/2 is specified as the autocorrelation function $\mathbf{R}$.

At each active learning step, the subsequent experimental design point is selected as the sample that minimizes the U-function over a population of $10^5$ randomly generated Monte Carlo samples. Also relying on training samples chosen according to U-function scoring metrics, active learning strategies that consider corrected variance predictions (Eq. \ref{eq:Var_LOOCV}) are additionally investigated.

The results are reported in terms of relative error between the predicted limit state function index, $\widehat{\beta_{g_j}}$, and the \emph{ground truth}, ${\beta_{g_j}}$, which is known in this setting. Formally, the error estimator $\varepsilon_{\beta_{g_j}}^{(i)}$, associated with a target limit state function $g_j$, is calculated for each conducted experiment $i$, as:
\begin{equation}
\varepsilon_{\beta_{g_j}}^{(i)}  =  \left| \frac{ \widehat{\beta}_{g_j}^{(i)} - \beta_{g_j} }{ \beta_{g_j}  } \right|.
\label{eq:error_metric}
\end{equation}

In total, 15 experiments are executed for each tested active learning strategy considering both U and U-LOO scoring function metrics: (i) training points selected with respect to target limit state function $g_1$, i.e., $ \mathcal{X}_{g_1}$, (ii) training points that specifically target limit state function $g_2$, i.e., $ \mathcal{X}_{g_2}$, (iii) alternate sequential active learning approach, i.e., $\mathcal{X}_{g_j}$, and (iv) training points selected based on the convergence criterion stated in Eq. \ref{eq:beta_strat}, i.e., $\mathcal{X}_{g^*_j}$. In order to objectively evaluate the error accumulated from both considered limit state functions, an error metric is formulated as: $\varepsilon_{\beta^{(i)}} =\sum_{j=1}^{m} \varepsilon_{\beta_{g_j}}^{(i)}$, providing a proxy for assessing the joint performance.

\begin{figure*}[h]
  \graphicspath{ {./figures/} }
  \centering
  \includegraphics{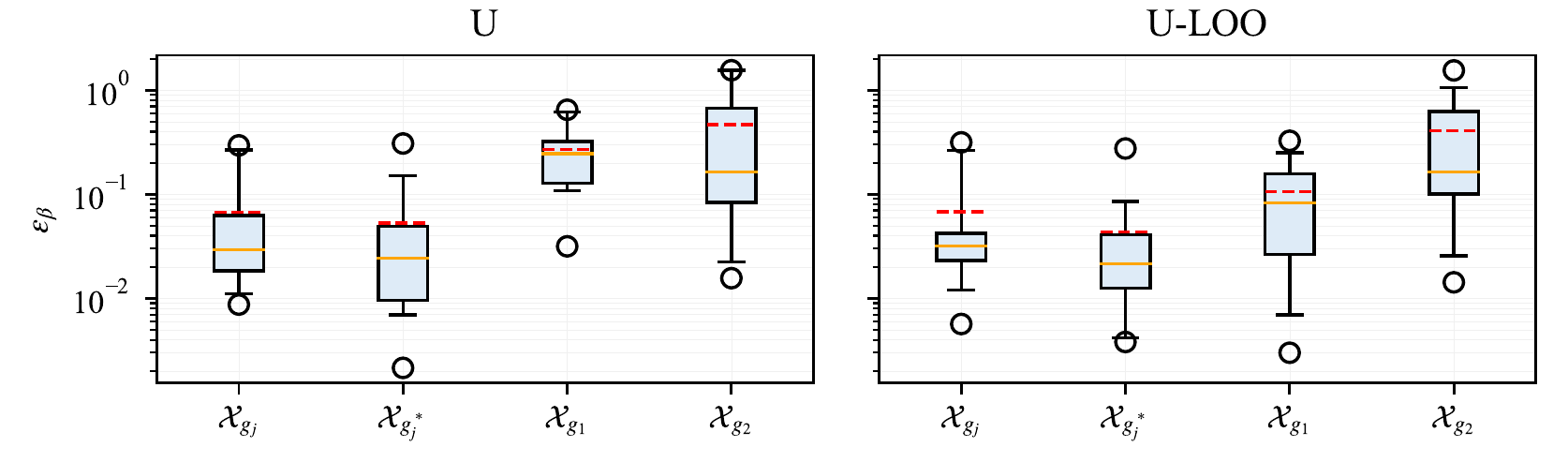}
  \caption{Box-plot representation of all investigated active learning approaches in terms of the combined relative error metric, $ \varepsilon_{\beta}$. (Left) U-function score metrics evaluated from PC-Kriging variance predictors. (Right) U-function score metrics calculated from corrected PC-Kriging variance predictors (U-LOO). \vspace{-0.1cm}}
  \label{Case1_BoxPlot}
\end{figure*}

\begin{table*}[h]
  \caption{Error estimator associated with a target limit state function, $\varepsilon_{\beta_{g_j}}$, and combined error estimator, $\varepsilon_{\beta}$.} \label{tab:label}
  \centering
    \begin{tabular}{|c|c|c|c|c|}
      \hline
         Strategy   & Learning metric & $ \mathbf{E} [\varepsilon_{\beta_{g_1}}](\sigma [\varepsilon_{\beta_{g_1}}] ) $ & $\mathbf{E} [\varepsilon_{\beta_{g_2}}](\sigma [\varepsilon_{\beta_{g_2}}] ) $ & $\mathbf{E} [\varepsilon_{\beta_{g}}](\sigma [\varepsilon_{\beta_{g}}] ) $ \\
\hline
\hline
$\mathcal{X}_{g_j}$ & U(PCK) & $3.7\cdot 10^{-2}( 7.1\cdot 10^{-2} )$ & $3.0\cdot 10^{-2} (6.2\cdot 10^{-2})$ & $6.7\cdot 10^{-2}(8.6\cdot 10^{-2}) $ \\
\hline
$\mathcal{X}_{g^*_j}$ & U(PCK) & $4.5\cdot 10^{-2}( 7.8\cdot 10^{-2} )$ & $8.2\cdot 10^{-3}(7.4\cdot 10^{-3})$ & $\mathbf{5.3\cdot 10^{-2}( 7.7\cdot 10^{-2}) }$ \\
\hline
$\mathcal{X}_{g_1}$ & U(PCK)  & $\mathbf{2.4\cdot 10^{-3}(1.8\cdot 10^{-3})}$ & $2.7\cdot 10^{-1}( 1.8\cdot 10^{-1})$ & $2.7\cdot 10^{-1}( 1.8\cdot 10^{-1})$ \\
\hline
$\mathcal{X}_{g_2}$ & U(PCK)  & $4.6\cdot 10^{-1}(6.0\cdot 10^{-1})$ & $\mathbf{7.1\cdot 10^{-3}(6.0\cdot 10^{-3}})$ & $4.7\cdot 10^{-1}(6.0\cdot 10^{-1}) $ \\
\hline
\hline
$\mathcal{X}_{g_j}$ &U(PCK-LOO)  & $5.8\cdot 10^{-2}(8.9\cdot 10^{-2})$ & $1.0\cdot 10^{-2}(7.2\cdot 10^{-3})$ & $6.8\cdot 10^{-2}(9.1\cdot 10^{-2}) $ \\
\hline
$\mathcal{X}_{g^*_j}$ & U(PCK-LOO) & $3.7\cdot 10^{-2}(6.6\cdot 10^{-2})$ & $\mathbf{6.0\cdot 10^{-3}(5.2\cdot 10^{-3})}$ & $\mathbf{4.3\cdot 10^{-2}(6.6\cdot 10^{-2} )}$ \\
\hline
$\mathcal{X}_{g_1}$  &U(PCK-LOO)  & $\mathbf{7.4\cdot 10^{-3}(1.6\cdot 10^{-2})}$ & $9.9\cdot 10^{-2}(9.5\cdot 10^{-2})$ & $1.1\cdot 10^{-1}(9.2\cdot 10^{-2}) $ \\
\hline
$\mathcal{X}_{g_2}$ & U(PCK-LOO)  & $4.0\cdot 10^{-1}(4.5\cdot 10^{-1})$ & $6.9\cdot 10^{-3}(4.3\cdot 10^{-3})$ & $4.1\cdot 10^{-1}(4.4\cdot 10^{-1}) $ \\
\hline
    \end{tabular}
    \label{table_Error_ULOO}
\end{table*}

\subsubsection*{Results and discussion}
Figure \ref{Case1: Pf_evolution} showcases the evolution of the surrogated event probabilities, $\mathrm{p}_{g_1}$ and $\mathrm{p}_{g_2}$, estimated from all tested active training strategies. In particular, the resulting expected probability, bounded by 30-60\% percentiles, is represented over 49 training samples. With the purpose of interpreting the influence of the predicted variance on the learning evolution, strategies resulting from U-based learning scores computed from both PCK’s variance (U) and corrected variance (U-LOO) are additionally compared.

As seen in the figure, PCK models trained with respect to a specific target limit state result inaccurate when the surrogate model is applied for the estimation of the other considered limit state function. Particularly, strategies $\mathcal{X}_{g_1}$ and $\mathcal{X}_{g_2}$ render inaccurate estimations and significant training variance across experiments for $\mathrm{p}_{g_2}$ and $\mathrm{p}_{g_1}$, respectively. In contrast, strategies that sequentially target both limit state functions within the available computational budget are able to yield a more balanced result in terms of accuracy and training stability. The straightforward alternate strategy, $\mathcal{X}_{g_j}$, for instance, concurrently reduces the inaccuracy gap resulting from both surrogated limit state functions over the training process. By implementing the strategy based on a converge-related metric, $\mathcal{X}_{g^*_j}$, more accurate predictions can be achieved for both limit state functions. This can be attributed to the fact that, at each learning step, the selected training point seeks the improvement of the most inaccurate surrogated limit state function.   

The performance reached at the end of the training process is additionally described in Table \ref{table_Error_ULOO}, listing the expected relative error metric and its corresponding standard deviation for each considered limit state function, $\varepsilon_{\beta_{g_j}}$, together with the total relative error, $\varepsilon_{\beta}$. For each listed error metric, the best strategy is highlighted. Evidently, the strategies $\mathcal{X}_{g_1}$ and $\mathcal{X}_{g_2}$ yield accurate predictions for their corresponding target limit state function, yet high relative errors are then observed when they are applied to the other limit state function, thus ultimately resulting in a global high error, $\varepsilon_{\beta}$, compared to the strategies $\mathcal{X}_{g_j}$ and $\mathcal{X}_{g^*_j}$. When considering U-LOO-based learning score metrics, $\mathcal{X}_{g^*_j}$ surprisingly becomes the best strategy for accurately representing the second limit state over all tested experiments, yet with a higher variability than $\mathcal{X}_{g_2}$. 

To further inspect the spread of the reported relative error metric $ \varepsilon_{\beta}$ over experiments, a box plot is shown in Figure \ref{Case1_BoxPlot}, delimiting the interquartile range, i.e., between 25\% and 75\% quantiles, and featuring a whisker that extends over 2.5\% and 97.5\%. Additionally, the mean and median values are indicated with solid orange and dashed red lines, respectively. In the figure, one can clearly observe that the strategy based on the converge-related metric, $\mathcal{X}_{g^*_j}$, yields a lower mean and median relative error in both U and U-LOO settings compared to its counterparts. While rendering accurate predictions for both limit state functions, the alternate active learning strategy, $\mathcal{X}_{g_j}$, logically results in higher variability over experiments compared to $\mathcal{X}_{g^*_j}$, as the target limit state function is not there intentionally assigned. With respect to the learning metric, the reported results also show that correcting the variance retrieved from the surrogate model leads to more accurate learning strategies. 

\subsection{Offshore wind substructure subject to ship collision accidental events}
In this second numerical experiment, we test the proposed active learning approaches for quantifying the probability associated with a repair and a failure event in the aftermath of an offshore wind substructure subject to ship collisions. All computations are conducted on an Intel Core $i9-10920X$ processor with a clock speed of 3.50 $GHz$. The assumed \emph{ground truth} corresponds, in this setting, to the substructure penetration, $\Delta_{p}$, computed from a simplified numerical model \citep{Timothee_phdthesis} based on limit plastic analysis. Since each numerical simulation requires approximately 3 minutes of computational time, a non-surrogated estimation of the failure and repair probabilities is computationally unfeasible. 

Both considered limit state functions are defined according to the resulting maximum substructure penetration, $\Delta_{p,max}$. A failure occurs if the maximum penetration exceeds a critical value, $\Delta_{F}$:
\begin{equation}
g_{F} \left(v_s, \rho_0\right) = \Delta_{F} - \Delta_{p,max}\left(v_s, \rho_0\right)  , 
\label{eq:gF_collision}
\end{equation}
whereas a repair event is stated as a function of a specified damage condition, $\Delta_{d}$, as:
\begin{equation}
g_{d} \left(v_s, \rho_0\right) =  \Delta_{d} - \Delta_{p,max}\left(v_s, \rho_0\right) , 
\label{eq:gR_collision}
\end{equation}
where $\Delta_{F}$ and $\Delta_{d}$ are specified as 3 and 2 meters, respectively. The random variables governing a collision scenario are the initial velocity of the ship, $v_s$, and the material flow stress, $\rho_0$, assumed here as elastic-perfectly plastic. These variables are probabilistically described as $v_s \backsim  \mathcal{N} \left(\mu = 3.0, \sigma = 0.6 \right)$ m/s and $\rho_0 \backsim  \mathcal{N} \left(\mu = 317, \sigma = 30 \right)$ N/mm$^2$, respectively. 
\begin{figure}[h!]
  \graphicspath{ {./figures/} }
  \centering
  \includegraphics{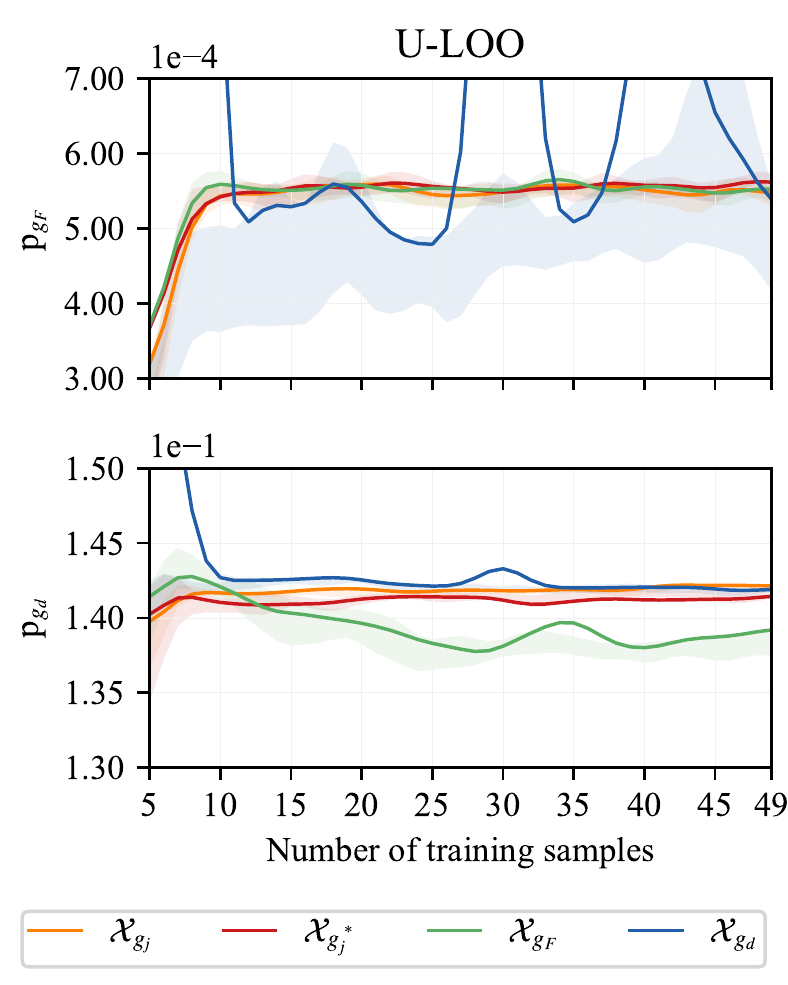}
  \caption{Active learning evolution in terms of the predicted probability associated with both considered events, $g_F$ and $g_d$, all relying on U-function metrics estimated from corrected variance predictors (U-LOO).}
  \label{Collision_evolution}
\end{figure}

\subsubsection*{Results and discussion}
The active learning evolution is showcased in Figure \ref{Collision_evolution}, representing both failure and damage probabilities over 10 experiments. Note that, in this case, only the corrected variance active learning scheme (U-LOO) is investigated. Since a closed-form solution of the analyzed failure and damage events is not available, all tested strategies are examined in terms of convergence. One can observe that with only a few training samples, most strategies easily reach convergence, yet active learning schemes that only target one specific limit state function (i.e., $\mathcal{X}_{g_F}$ and $\mathcal{X}_{g_d}$), are evidently more unstable when applied to the other considered limit state. Instead, active learning strategies that interchangeably target all considered limit state functions converge to both estimated failure and damage event probabilities, all trained under the same computational budget.

\section{CONCLUSIONS}
Surrogate-based active learning strategies for reliability analysis effectively yield accurate predictions for a specifically targeted limit state function, yet they may render inaccurate estimates for other regions within the design space. This paper reveals that, by sequentially targeting multiple limit state functions throughout the training process, combined active learning strategies are able to achieve a balanced computational budget allocation, resulting in low overall prediction errors. 




\section*{Acknowledgement}
Mr. Morán gratefully acknowledges the support received
by the National Fund for Scientific Research
in Belgium F.R.I.A. - F.N.R.S.

\bibliographystyle{elsarticle-num} 
\bibliography{bib}

\begin{thebibliography}{1}
\expandafter\ifx\csname url\endcsname\relax
  \def\url#1{\texttt{#1}}\fi
\expandafter\ifx\csname urlprefix\endcsname\relax\def\urlprefix{URL }\fi
\expandafter\ifx\csname href\endcsname\relax
  \def\href#1#2{#2} \def\path#1{#1}\fi

\bibitem{MORATO2022102140}
P.~G. Morato, K.~G. Papakonstantinou, C.~P. Andriotis, J.~S. Nielsen, P.~Rigo,
  \href{https://www.sciencedirect.com/science/article/pii/S0167473021000631}{{Optimal
  inspection and maintenance planning for deteriorating structural components
  through dynamic Bayesian networks and Markov decision processes}}, Structural
  Safety 94 (2022) 102140.
\newblock \href
  {https://doi.org/https://doi.org/10.1016/j.strusafe.2021.102140}
  {\path{doi:https://doi.org/10.1016/j.strusafe.2021.102140}}.

\bibitem{Schobi_PCKriging}
R.~Schobi, B.~Sudret, J.~Wiart, {POLYNOMIAL-CHAOS-BASED KRIGING}, International
  Journal for Uncertainty Quantification 5~(2) (2015) 171--193.

\bibitem{doi:10.2514/1.34321}
B.~J. Bichon, M.~S. Eldred, L.~P. Swiler, S.~Mahadevan, J.~M. McFarland,
  \href{https://doi.org/10.2514/1.34321}{Efficient global reliability analysis
  for nonlinear implicit performance functions}, AIAA Journal 46~(10) (2008)
  2459--2468.
\newblock \href {http://arxiv.org/abs/https://doi.org/10.2514/1.34321}
  {\path{arXiv:https://doi.org/10.2514/1.34321}}, \href
  {https://doi.org/10.2514/1.34321} {\path{doi:10.2514/1.34321}}.

\bibitem{ECHARD2011145}
B.~Echard, N.~Gayton, M.~Lemaire,
  \href{https://www.sciencedirect.com/science/article/pii/S0167473011000038}{{AK-MCS:
  An active learning reliability method combining Kriging and Monte Carlo
  Simulation}}, Structural Safety 33~(2) (2011) 145--154.
\newblock \href
  {https://doi.org/https://doi.org/10.1016/j.strusafe.2011.01.002}
  {\path{doi:https://doi.org/10.1016/j.strusafe.2011.01.002}}.

\bibitem{MOUSTAPHA2022102174}
M.~Moustapha, S.~Marelli, B.~Sudret,
  \href{https://www.sciencedirect.com/science/article/pii/S0167473021000965}{Active
  learning for structural reliability: Survey, general framework and
  benchmark}, Structural Safety 96 (2022) 102174.
\newblock \href
  {https://doi.org/https://doi.org/10.1016/j.strusafe.2021.102174}
  {\path{doi:https://doi.org/10.1016/j.strusafe.2021.102174}}.

\bibitem{XIAO2018404}
N.-C. Xiao, M.~J. Zuo, W.~Guo,
  \href{https://www.sciencedirect.com/science/article/pii/S0307904X18300829}{Efficient
  reliability analysis based on adaptive sequential sampling design and
  cross-validation}, Applied Mathematical Modelling 58 (2018) 404--420.
\newblock \href {https://doi.org/https://doi.org/10.1016/j.apm.2018.02.012}
  {\path{doi:https://doi.org/10.1016/j.apm.2018.02.012}}.

\bibitem{le2015cokriging}
L.~Le~Gratiet, C.~Cannamela, Cokriging-based sequential design strategies using
  fast cross-validation techniques for multi-fidelity computer codes,
  Technometrics 57~(3) (2015) 418--427.

\bibitem{Timothee_phdthesis}
T.~Pire, Crashworthiness of offshore wind turbine jackets based on the
  continuous element method, {PhD} dissertation, University of Liège, Belgium
  (2018).

\end{thebibliography}

\clearpage

\end{document}